\documentclass[final]{cvpr}

\usepackage{times}
\usepackage{epsfig}
\usepackage{graphicx}
\usepackage{amsmath}
\usepackage{amssymb}

\usepackage{ntheorem}
\usepackage{wrapfig,graphicx}
\usepackage{thmtools}
\usepackage{booktabs}       
\newcommand{\MC}{\mathcal}
\newcommand{\MBB}{\mathbb}
\newcommand{\MBF}{\mathbf}
\newtheorem{definition}{Definition}
\newtheorem{theorem}{Theorem}

\newtheorem{proposition}{Proposition}

\newtheorem{remark}{Remark}

\usepackage[pagebackref=true,breaklinks=true,colorlinks,bookmarks=false]{hyperref}

\def\cvprPaperID{3520} 
\def\confYear{CVPR 2021}

\begin{document}
\title{Conditional Bures Metric for Domain Adaptation}

\author{You-Wei Luo\textsuperscript{1} ~~~ Chuan-Xian Ren\textsuperscript{1,2}\thanks{Corresponding Author.} ~~\\
\textsuperscript{1}School of Mathematics, Sun Yat-Sen University, China\\
\textsuperscript{2}Pazhou Lab, Guangzhou, China\\
{\tt\small luoyw28@mail2.sysu.edu.cn,  rchuanx@mail.sysu.edu.cn}
}
\maketitle

\begin{abstract}
   As a vital problem in classification-oriented transfer, unsupervised domain adaptation (UDA) has attracted widespread attention in recent years. Previous UDA methods assume the marginal distributions of different domains are shifted while ignoring the discriminant information in the label distributions. This leads to classification performance degeneration in real applications. In this work, we focus on the conditional distribution shift problem which is of great concern to current conditional invariant models. We aim to seek a kernel covariance embedding for conditional distribution which remains yet unexplored. Theoretically, we propose the Conditional Kernel Bures (CKB) metric for characterizing conditional distribution discrepancy, and derive an empirical estimation for the CKB metric without introducing the implicit kernel feature map. It provides an interpretable approach to understand the knowledge transfer mechanism. The established consistency theory of the empirical estimation provides a theoretical guarantee for convergence. A conditional distribution matching network is proposed to learn the conditional invariant and discriminative features for UDA. Extensive experiments and analysis show the superiority of our proposed model.
\end{abstract}

\thispagestyle{empty}
\pagestyle{empty}

\section{Introduction}
Large-scale data with sufficient annotations are vital sources of machine learning. However, the data collected from the real-world scenarios are usually unlabeled and the manual annotations are expensive. Recent advances in transfer learning yields plenty of methods for dealing with the shortage of labeled data. These methods aim to transfer the knowledge on a labeled source domain to a target domain with few or no annotations, such setting is also known as domain adaptation \cite{pan2009survey}.

\begin{figure}
\centering
\includegraphics[width=0.475\textwidth]{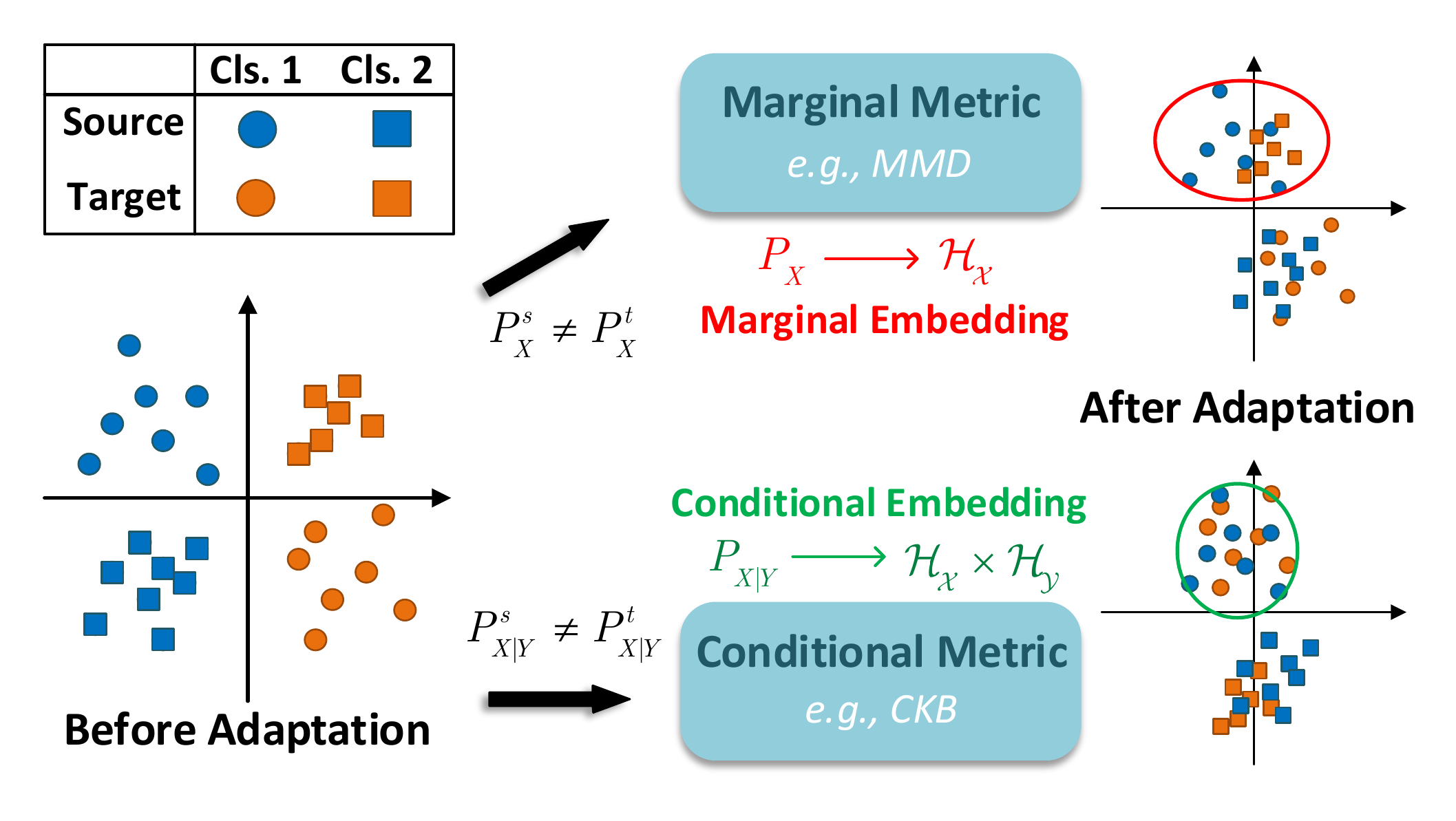}
\caption{Illustration of the conditional shift problem. Previous metrics that only consider the marginal distribution discrepancy may lead to a misaligned conditional distribution, \ie, the red circle region. On the bottom, the class-level alignment is achieved by exploiting the conditional distribution embedding metric.}\label{fig:Problem_Illustration}
\vspace{-10pt}
\end{figure}

The most common assumption in Unsupervised Domain Adaptation (UDA) is that the labeled source domain and unlabeled target domain have the same feature spaces, but different marginal distributions \cite{pan2009survey}, \ie, $\MC{X}^s=\MC{X}^t$, $P_X^s\neq P_X^t$. This assumption is also called covariate shift \cite{shimodaira2000improving} and sample selection bias \cite{zadrozny2004learning}. Ben-David \etal \cite{ben2007analysis} give a theoretical insight into the domain adaptation problem, they show that the risk of the target domain is mainly bounded by the risk of the source domain and the discrepancy between distributions of two domains. Inspired by this theory, many methods are proposed to mitigate the discrepancy between feature distributions of the source and target domains, \eg, explicit discrepancy minimization via Maximum Mean Discrepancy (MMD) \cite{gretton2012kernel,long2015learning}, domain invariant feature learning \cite{pan2010domain}, Optimal Transport (OT) based feature matching \cite{courty2016optimal,li2020Enhanced,zhang2019optimal}, manifold based feature alignment \cite{gong2012geodesic}, statistical moment matching \cite{long2015learning,sun2016return} and adversarial domain adaptation \cite{ganin2016domain}. These methods are proved to be effective in minimizing the marginal discrepancy and alleviating the domain shift problem. However, this assumption may lead to the omission of discriminant information in the label distributions, which is described in Figure \ref{fig:Problem_Illustration}. Recent advancements \cite{li2020maximum,long2018conditional,luo2020unsupervised} show that the adaptation models will be more discriminative on the target domain if the target label information (\eg, pseudo labels) is explored carefully.

Extended from the marginal shift assumption, the conditional shift problem is studied to build a conditional invariant model \cite{zhang2013domain}, \ie, $P_{X|Y}^s=P_{X|Y}^t$. The most critical problem is to construct a framework which can explicitly reflect the relation between different conditional distributions. Zhao \etal \cite{zhao2019learning} prove a new generalization bound which quantitatively reflects the underlying structure of the conditional shift problem. Several works have also been made in the field of conditional/joint distribution matching for domain adaptation, \eg, multi-layer feature approximation \cite{Long2017Deep}, conditional variants of MMD \cite{kang2020contrastive,li2020maximum,zhu2020deep}, conditional invariant learning with causal interpretations \cite{gong2016domain,ren2018generalized}, OT based joint distribution models \cite{bhushan2018deepjdot,courty2017joint}.



In this paper, we aim to estimate the transport cost in Reproducing Kernel Hilbert Space (RKHS) for the continuous conditional distributions. Inspired by pioneering work \cite{fukumizu2009kernel}, which employs the conditional covariance operator on the RKHS to characterize the independence, we define transport cost estimation on the set of conditional covariance operators called Conditional Kernel Bures (CKB) metric. By virtue of the conditional covariance operator and OT theory, we prove that the CKB metric reflects the discrepancy between two conditional distributions directly. This result can be taken as an extension of the marginal distribution embedding property in MMD \cite{gretton2012kernel} and kernel Bures metric \cite{zhang2019optimal}. An explicit empirical estimation of the CKB metric and its consistency theory are presented. Further, we apply it to the proposed conditional distribution matching network. Extensive experiment results show the effectiveness of the CKB metric and the superiority of the proposed model. Our contributions are summarized as follows.
\begin{itemize}
\item A novel CKB metric for characterizing conditional distribution discrepancy is proposed, and the kernel embedding property of the CKB metric is proved to show that it is well-defined on conditional distributions. This metric is also exactly the OT between conditional distributions, which provides an interpretable approach to understand the knowledge transfer mechanism.
\item An explicit empirical estimation of the CKB metric is derived, which provides a computable measurement for conditional domain discrepancy. The asymptotic property of the estimation is proved which provides a rigorous theoretical guarantee for convergence.
\item A conditional distribution matching network based on the CKB metric is proposed for discriminative domain alignment, and a joint distribution matching variant is further extended. The SOTA results in extensive experiments validate the model's effectiveness.
\end{itemize}

\section{Related Work}
\textbf{Unsupervised Domain Adaptation.} Based on the distribution shift assumption, the UDA methods can be roughly categorized as follows. Domain invariant feature learning methods like Transfer Component Analysis (TCA) \cite{pan2010domain} try to learn a set of transfer components that make the corresponding distribution robust to the change of domains. OT based methods mitigate the domain discrepancy by minimizing the cost of transporting the source samples to the target domain. It has been shown that OT alignment is equivalent to minimizing the KL divergence \cite{courty2016optimal} or Wasserstein distance \cite{zhang2019optimal} between the distributions. Moment matching methods attempt to minimize the distribution discrepancy via statistical moments, \eg, Domain Adaptation Network (DAN) \cite{long2015learning} for the first order matching and CORAL \cite{sun2016return} the second order. Manifold alignment methods take the domains as the points on the manifold and align the domains under the manifold metric \cite{gong2012geodesic,luo2020unsupervised}. Adversarial based methods \cite{ganin2016domain,tang2020discriminative} alternatively optimize the feature generator and domain discriminator, which are respectively supposed to be domain-confusable and discriminative, to achieve domain confusion. Extended from the marginal distribution assumption, recent works \cite{bhushan2018deepjdot,courty2017joint,li2020Enhanced,long2018conditional,Long2017Deep} show that the models yield promising results by introducing the label information. Joint Adaptation Network (JAN) \cite{Long2017Deep} builds a joint distribution alignment model via the features from different hidden layers. Conditional Domain Adversarial Network (CDAN) \cite{long2018conditional} extends the Domain Adversarial Neural
Network (DANN) \cite{ganin2016domain} by exploring a multilinear map to describe the conditional variables in adversarial training.

\textbf{Optimal Transport.} Recently, OT has been successively applied to the UDA problem \cite{bhushan2018deepjdot,courty2017joint,courty2016optimal,li2020Enhanced,zhang2019optimal}. Courty \etal \cite{courty2016optimal} deal with UDA based on the Kantorovitch formulation of OT, which allows to define the well-known Wasserstein distance between the domain distributions. As a variant of Wasserstein distance, Bures metric has been of great interest to various research fields like quantum information, information theory and Riemannian geometry \cite{bhatia2019bures}. The original Bures metric is defined on the set of Positive Semi-Definite (PSD) matrices and cannot be used to measure the distribution discrepancy. In \cite{zhang2019optimal}, Zhang \etal extend the OT problem to RKHS, and then define the kernel Wasserstein distance and kernel Bures metric. They show the covariance embedding in RKHS is injective which implies that the kernel Bures metric defines a metric on the distributions. However, these discrepancy measures mainly focus on the marginal distribution. To exploit the label information, joint distribution OT models \cite{bhushan2018deepjdot,courty2017joint} seek an optimal joint transport map that minimizes the generalized cost associated to the joint space of features and labels $X\times Y$. Enhanced Transport Distance (ETD) \cite{li2020Enhanced} uses the prediction feedback from the classifier to reweigh the transport cost. Differing from the above OT based methods which are formulated on discrete joint distribution or marginal distribution, our work focuses on the explicit estimation of OT between conditional distributions under the continuous case.



\section{OT for Conditional Distribution}
In this section, we first review the definitions and properties of conditional covariance operator and Kantorovitch's OT in RKHS, which are the fundamentals of the proposed CKB metric. Then we present the theoretical definition and property of the CKB metric. Finally, we provide the empirical estimation and its asymptotic property.
\subsection{Preliminary}

\textbf{Conditional Covariance Operators. }
Let $(\MC{X},\MC{B})$ be a measure space with Borel $\sigma$-field $\MC{B}$. Denote $(\MC{H}_{\MC{X}},k_{\MC{X}})$ as the RKHSs of $\MC{X}$, which is generated by the positive definite kernels $k_{\MC{X}}$. The mean element $\mu_{X}$ in $\MC{H}_\MC{X}$ with law $P_X$ is given by $\mu_{X} = \MBB{E}_X \left[ \phi(X) \right]$, where $\phi$ is the nonlinear feature map of $\MC{H}_{\MC{X}}$. It is assumed that $\phi(x)=k_{\MC{X}}(x,\cdot)$ satisfies the reproducing properties $\left<\phi(x),\phi(x')\right>_{\MC{H}_{\MC{X}}}=k_{\MC{X}}(x,x')$ and $\left<\phi(x),f\right>_{\MC{H}_{\MC{X}}} = f(x)$, $\forall f \in \MC{H}_{\MC{X}}$.

To explore the casual connection between $\MC{X}$ and $\MC{Y}$, we consider the pair $(X,Y):\Omega \to \MC{X} \times \MC{Y}$ with probability measure $P_{XY}\in \text{Pr}(\MC{X},\MC{Y})$, where $\text{Pr}(\MC{X},\MC{Y})$ is the set of Borel probability measures on $(\MC{X},\MC{Y})$. Given a joint measure $( \MC{H}_{\MC{X}} \times \MC{H}_{\MC{Y}}, \MC{B}_{\MC{X}} \times \MC{B}_{\MC{Y}} )$, its corresponding \textit{cross-covariance operator} \cite{baker1973joint} $\MBF{R}_{XY}: \MC{H}_{\MC{Y}} \to \MC{H}_{\MC{X}}$ satisfies that $\forall ~ f \in \MC{H}_{\MC{X}},~g \in \MC{H}_{\MC{Y}},$
\begin{equation*}\label{eq:cross-covariance_define}
\left< f, \MBF{R}_{XY}g \right>_{\MC{H}_{\MC{X}}} = \MBB{E}_{XY} \left[ f(X)g(Y) \right] - \MBB{E}_X \left[ f(X) \right] \MBB{E}_Y \left[ g(Y) \right]
\end{equation*}
Formally, $\MBF{R}_{XY}$ is defined as \cite{song2009hilbert}
\begin{equation*}\label{eq:cross-covariance_compute}
\MBF{R}_{XY}=\MBB{E}_{XY} \left[ \left( \phi(X) - \mu_{X} \right) \otimes \left( \psi(Y) - \mu_{Y} \right) \right].
\end{equation*}
If $Y$ equals to $X$, $\MBF{R}_{XX}$ is just the covariance operator on $\MC{H}_{\MC{X}}$. Based on the \textit{cross-covariance operator}, we further consider the conditional covariance of $\phi (X)$ w.r.t. the conditioning variable $Y$. The \textit{conditional covariance operator} $\MBF{R}_{XX|Y}$ is usually written as \cite{fukumizu2009kernel}
\begin{equation*}\label{eq:conditional-covariance_compute}
\MBF{R}_{XX|Y} = \MBF{R}_{XX} - \MBF{R}_{XY} \MBF{R}_{YY}^{-1} \MBF{R}_{YX}.
\end{equation*}
Note that $\MBF{R}_{YY}$ may be non-invertible, especially in the real-world applications with finite samples. When necessary conditions are fulfilled \cite{fukumizu2009kernel}, the conditional covariance operator also satisfies that
\begin{equation*}\label{eq:conditional-covariance_property}
\left< f, \MBF{R}_{XX|Y} f \right>_{\MC{H}_{\MC{Y}}} = \MBB{E}_Y \left[ \text{Var}_{X|Y}[f(X)|Y] \right], ~~~~ \forall~ f \in \MC{H}_{\MC{X}}.
\end{equation*}

\textbf{Kantorovitch's OT in RKHS. }
For any two distributions $P^{s}_{X}, P^{t}_{X} \in \text{Pr}(\MC{X})$, let $\Pi(P^{s}_{X} \times P^{t}_{X})$ be the set of probabilistic couplings, the Kantorovitch formulation of OT is
\begin{equation}\label{eq:Kantorvitch}
\gamma^* = \inf_{\gamma \in \Pi(P^{s}_{X} \times P^{t}_{X})} \int_{\MC{X} \times \MC{Y}} d^2(\MBF{x}^s,\MBF{x}^t) d \gamma(\MBF{x}^s,\MBF{x}^t).
\end{equation}
The Kantorovitch problem in Eq.~\eqref{eq:Kantorvitch} is also equivalent to the Wasserstein distance. Under the Gaussian measures, if the distributions $P^{s}_{X}$ and $P^{t}_{X}$ have the same expectations, the Wasserstein distance between them is equivalent to the Bures metric between their covariance matrices. Let $\MBB{S}^+(d)$ be the set of $d \times d$ PSD matrices; for any PSD matrix $\MBF{\Sigma}$, its unique square root $\sqrt{\MBF{\Sigma}}$ is defined by $\MBF{\Sigma}=\sqrt{\MBF{\Sigma}} \sqrt{\MBF{\Sigma}}$. The Bures metric is defined by
\begin{equation*}\label{eq:bures_define}
d_{\text{B}}^2(\MBF{\Sigma}^{s}_{XX},\MBF{\Sigma}^{t}_{XX}) = \text{tr} \left( \MBF{\Sigma}^{s}_{XX} + \MBF{\Sigma}^{t}_{XX} - 2\MBF{\Sigma}^{st}_{XX} ~\right),
\end{equation*}
where $\MBF{\Sigma}^{st}_{XX}=\sqrt{ \sqrt{\MBF{\Sigma}^{s}_{XX}} \MBF{\Sigma}^{t}_{XX} \sqrt{\MBF{\Sigma}^{s}_{XX}} }$ and $\MBF{\Sigma}^{s}_{XX}$ and $\MBF{\Sigma}^{t}_{XX}$ are the covariance matrices of $P^{s}_{X}$ and $P^{t}_{X}$, respectively. Recent work shows that the Bures metric is also related to the Riemannian geometry, as it can be taken as the metric on PSD manifold \cite{bhatia2019bures}. Though the Bures metric defines a metric on $\MBB{S}^+(d)$, it cannot reflect discrepancy between distributions $P^{s}_{X}$ and $P^{t}_{X}$.

The kernel Bures metric \cite{zhang2019optimal} generalizes the PSD setting in Bures metric to the infinite-dimensional RKHS $\MC{H}$. Let $\MBB{S}^+(\MC{H}_{\MC{X}}) \subseteq \MC{H}_{\MC{X}} \times \MC{H}_{\MC{X}}$ be the set of all positive, self-adjoint, and trace-class operators on $\MC{H}_{\MC{X}}$ with kernel $k_{\MC{X}}$, the kernel Bures metric $d_{\text{KB}}(\cdot,\cdot)$ on $\MBB{S}^+(\MC{H}_{\MC{X}})$ is written as:
\begin{equation*}\label{eq:kernel-bures_define}
d_{\text{KB}}^2(\MBF{R}^{s}_{XX},\MBF{R}^{t}_{XX}) = \text{tr} \left( \MBF{R}^{s}_{XX} + \MBF{R}^{t}_{XX} - 2\MBF{R}^{st}_{XX} ~\right),
\end{equation*}
where $\MBF{R}^{st}_{XX} = \sqrt{ \sqrt{\MBF{R}^{s}_{XX}} \MBF{R}^{t}_{XX} \sqrt{\MBF{R}^{s}_{XX}} }$ and $\MBF{R}^{s}_{XX}, \MBF{R}^{t}_{XX}$ are the covariance operators of $P^{s}_{X}$ and $P^{t}_{X}$ on $\MC{H}_{\MC{X}}$, respectively. Note the kernel Bures is exactly the transport cost in RKHS when the push-forward measures $\phi \# P^{s}_{X}$ and $\phi \# P^{t}_{X}$ are Gaussian \cite{zhang2019optimal}. Zhang \etal \cite{zhang2019optimal} prove that if the measurable space $(\MC{X},\MC{B}_{\MC{X}})$ is locally compact and Hausdorff, the embedding $P^{s}_{X} \mapsto \MBF{R}^{s}_{XX}, \forall P^{s}_{X} \in \text{Pr}(\MC{X})$ is injective. It turns out that $d_{\text{KB}}(\cdot,\cdot)$ defines a metric on $\text{Pr}(\MC{X})$, which no longer holds for the Bures metric. With this property, the kernel Bures metric can be used to quantify the discrepancy between two distributions.

\subsection{Conditional Kernel Bures Metric}
To introduce conditional distribution to OT, we develop the kernel covariance embedding property for conditional distributions and apply it to the kernel Bures metric. The CKB metric for conditional distributions is now defined.

\begin{definition}\label{def:conditional-kernel-bures}
The Conditional Kernel Bures (CKB) metric between two conditional distributions $P^{s}_{X|Y}, P^{t}_{X|Y} \in \rm{Pr}(\MC{X}|\MC{Y})$ is defined as
\begin{small}
\begin{equation}\label{eq:conditional-kernel-bures_define}
 d_{\emph{CKB}}^2(\MBF{R}^{s}_{XX|Y},\MBF{R}^{t}_{XX|Y}) =  \emph{tr} \left( \MBF{R}^{s}_{XX|Y} + \MBF{R}^{t}_{XX|Y} - 2\MBF{R}^{st}_{XX|Y} ~\right),
\end{equation}
\end{small}
where
\begin{small}
$\MBF{R}^{st}_{XX|Y}=\sqrt{ \sqrt{\MBF{R}^{s}_{XX|Y}} \MBF{R}^{t}_{XX|Y} \sqrt{\MBF{R}^{s}_{XX|Y}} }$
\end{small}
.
\end{definition}
\begin{proposition}\label{prop:CKB_define_metric}
CKB $d_{\emph{CKB}}(\cdot,\cdot)$ defines a metric on $\MBB{S}^+(\MC{H}_{\MC{X}})$.
\end{proposition}

Recall that the conditional covariance operator $\MBF{R}_{XX|Y}$ is also positive, self-adjoint, and trace-class on $\MC{H}_{\MC{X}}$ \cite{fukumizu2009kernel}. Thus, we can deduce from Proposition \ref{prop:CKB_define_metric} that the CKB metric is well-defined on the conditional covariance operators.

The injective property of mean embedding $\MBB{E}_{X} \left[\phi(X)\right]$ \cite{gretton2012kernel} and covariance embedding $\MBF{R}_{XX}$ \cite{zhang2019optimal} in RKHS give the theoretical insights into how two distributions are matched via the defined metrics, \eg, MMD and kernel Bures metric. Similarly, we also make connection between the CKB metric and conditional distributions. Note that though the above embedding properties are well studied, they only consider connections between the operators and the marginal distributions. As the embedding property between the covariance operators and conditional distributions is unexplored, our work focuses on extending the CKB metric to a metric on conditional distributions ${\rm Pr}(\MC{X}|\MC{Y})$. For convenience, we denote the set of measures that satisfy the 3-splitting property \cite{zhang2019optimal} by ${\rm Pr}^s(\MC{X}|\MC{Y}=y)$ and the direct sum by $\MC{H}_{\MC{X}}\oplus \MC{H}_{\MC{Y}}$.


\begin{theorem}\label{thm:conditional-kernel-bures_metricThm}
Let $(\MC{X},\MC{B}_{\MC{X}})$ be the locally compact and Hausdorff measurable space and $k$ be $c_0$-universal kernel. Assuming that $\left( \phi (X),\psi (Y) \right)$ is a Gaussian random variable in $\MC{H}_{\MC{X}}\oplus \MC{H}_{\MC{Y}}$. 
For any $P^{s}_{X|Y}, P^{t}_{X|Y} \in {\rm Pr}^s(\MC{X}|\MC{Y})$, we have
\begin{equation*}\label{eq:conditional-kernel-bures_metricThm}
d_{\emph{CKB}}(\MBF{R}^{s}_{XX|Y},\MBF{R}^{t}_{XX|Y})=0 ~~~~ \Longrightarrow ~~~~ P^{s}_{X|Y}=P^{t}_{X|Y}.
\end{equation*}
\end{theorem}

The above theorem shows that the CKB metric $d_{\text{CKB}}(\cdot,\cdot)$ defines a metric on $\text{Pr}(\MC{X|Y})$ if some conditions are satisfied. Note that the CKB metric is exactly the minimized OT cost between two conditional distributions since $\phi \# P^s_{(X,Y)}$ and $\phi \# P^t_{(X,Y)}$ are also Gaussian. Thus, it can be used to measure the discrepancy between two conditional distributions. The condition $c_0$-universal \cite{sriperumbudur2011universality} in Theorem \ref{thm:conditional-kernel-bures_metricThm} is satisfied by many common kernels, \eg, Gaussian kernel and Laplacian kernel. The assumption of Gaussian random variable can be taken as the extension of Gaussian distribution which takes values in RKHS \cite{klebanov2020rigorous}. Recall that the feature maps $\phi(\cdot)$ and $\psi(\cdot)$ are implicit, so the conditional covariance operator $\MBF{R}_{XX|Y}$ is not formulable in practical computation of the CKB metric. To present an explicit formulation of the CKB metric, we use the kernel trick, \ie, $\left<\phi(x),\phi(x')\right>_{\MC{H}_{\MC{X}}}=k_{\MC{X}}(x,x')$, to avoid the explicit nonlinear maps in the next section.

\subsection{Empirical Estimation of the Conditional Kernel Bures Metric}\label{subsec:empirical_estimation_ckb}

Let $\MC{D}^{s} = \{ (\MBF{x}^s_i,\MBF{y}^s_i) \}_{i=1}^{n}$ and $\MC{D}^{t} = \{ (\MBF{x}^t_j,\MBF{y}^t_j) \}_{j=1}^{m}$ be two sets of samples, which are assumed to be drawn i.i.d. from $P^s_{XY}$ and $P^t_{XY}$, respectively. Note that $x^{s/t}_i \in \MBB{R}^{d}, y^{s/t}_i \in \MBB{R}^{c}$, and we map the data $x^{s/t}_i$ (resp. $y^{s/t}_i$) to the RKHS $\MC{H}_{\MC{X}}$ (resp. $\MC{H}_{\MC{Y}}$) with the implicit feature map $\phi$ (resp. $\psi$). Let $\MBF{K}^{s/t}_{XX}$, $\MBF{K}^{s/t}_{YY}$ and $\MBF{K}^{ts}_{XX}$ be the explicit kernel matrices computed as $(\MBF{K}^{s/t}_{XX})_{ij} = k_{\MC{X}}(\MBF{x}_i^{s/t},\MBF{x}_j^{s/t})$, $(\MBF{K}^{s/t}_{YY})_{ij} = k_{\MC{Y}}(\MBF{y}_i^{s/t},\MBF{y}_j^{s/t})$ and $(\MBF{K}^{ts}_{XX})_{ij} = k_{\MC{X}}(\MBF{x}_i^{t},\MBF{x}_j^{s})$, respectively. Denote the feature map matrices by $\MBF{\Phi}_{s/t}$ and $\MBF{\Psi}_{s/t}$.
Their cross-covariance matrices can be written as $\hat{\MBF{R}}^{s}_{XY} = \frac{1}{n} \MBF{\Phi}_{s} \MBF{H}_{n} \MBF{\Psi}_{s}^T , ~ \hat{\MBF{R}}^{t}_{XY} = \frac{1}{m} \MBF{\Phi}_{t} \MBF{H}_{m} \MBF{\Psi}_{t}^T$, where $\MBF{H}_{n} = \MBF{I}_n - \frac{1}{n} \MBF{1}_n \MBF{1}_n^T$ is the $n\times n$ centering matrix, and $\MBF{1}_n$ is $n$-dimensional vector with all elements equal to 1. As the covariance matrix $\hat{\MBF{R}}^{s/t}_{YY}$ is always rank-deficient under the finite sample case, we regularize it as 
\begin{equation}\label{eq:conditional-covariance_regular}
\hat{\MBF{R}}_{XX|Y} = \hat{\MBF{R}}_{XX} - \hat{\MBF{R}}_{XY} \left( \hat{\MBF{R}}_{YY} + \varepsilon \MBF{I} \right)^{-1} \hat{\MBF{R}}_{YX},
\end{equation}
where $\varepsilon > 0$ is the regularization parameter. Denote the matrices
\begin{equation*}\label{eq:matrix-B}
\begin{split}
\MBF{B}_{s} \triangleq & ~ \MBF{I}_{n} - \frac{1}{n\varepsilon} \left[ \MBF{G}^s_{Y} - \MBF{G}^s_{Y} \left( \MBF{G}^s_{Y} + \varepsilon n \MBF{I}_n \right)^{-1}\MBF{G}^s_{Y} \right], \\
\MBF{B}_{t} \triangleq & ~ \MBF{I}_{m} - \frac{1}{m\varepsilon} \left[ \MBF{G}^t_{Y} - \MBF{G}^t_{Y} \left( \MBF{G}^t_{Y} + \varepsilon m \MBF{I}_m \right)^{-1}\MBF{G}^t_{Y} \right],
\end{split}
\end{equation*}
where
\begin{equation*}\label{eq:matrix-Gram-G}
\MBF{G}^{s}_{X/Y} = \MBF{H}_{n} \MBF{K}^{s}_{XX/YY} \MBF{H}_{n}, \quad \MBF{G}^{t}_{X/Y} = \MBF{H}_{m} \MBF{K}^{t}_{XX/YY} \MBF{H}_{m}
\end{equation*}
are the centralized kernel matrices. With the decomposition $\MBF{B}_{s/t}=\MBF{C}_{s/t} \MBF{C}_{s/t}^T$, the conditional covariance operator $\hat{\MBF{R}}^s_{XX|Y}$ can be reformulated as ($\hat{\MBF{R}}^t_{XX|Y}$ is the same)
\begin{equation}\label{eq:conditional-covariance_reformulation}
\hat{\MBF{R}}^s_{XX|Y}=\frac{1}{n} \MBF{\Phi}_s \MBF{H}_{n} \MBF{C}_s \left( \MBF{\Phi}_s \MBF{H}_{n} \MBF{C}_s \right)^T.
\end{equation}

\begin{proposition}\label{prop:positive-B}
If $k_{\MC{Y}}$ is positive definite kernel, then $\MBF{B}_{s}$ and $\MBF{B}_{t}$ are positive definite for any $\varepsilon>0$. Especially, we have
\begin{equation*}\label{eq:reformulated-B}
\MBF{B}_{s} = \varepsilon n \left( \MBF{G}^s_{Y} + \varepsilon n \MBF{I}_n \right)^{-1}, ~~~ \MBF{B}_{t} = \varepsilon m \left( \MBF{G}^t_{Y} + \varepsilon m \MBF{I}_m \right)^{-1}.
\end{equation*}
\end{proposition}

\begin{remark}\label{rem:empirical-estimation-CKB}
Proposition~\ref{prop:positive-B} shows that $\MBF{B}_{s/t}$ is positive definite with a positive definite kernel $k_{\MC{Y}}$ (\eg, Gaussian kernel and Laplacian kernel), so the decomposition $\MBF{B}_{s/t} = \MBF{C}_{s/t} \MBF{C}_{s/t}^T$ always exists. But, such a decomposition is not unique, \eg, Cholesky factorization and eigendecomposition. Here we compute $\MBF{C}_{s}$ based on the Eigenvalue Decomposition (EVD) of $\MBF{B}_{s}$ as ($\MBF{C}_{t}$ is the same)
\begin{equation*}\label{eq:Eigen-Matrix-B}
\MBF{B}_{s} = \MBF{U}_{s} \MBF{D}_{s} \MBF{U}_{s}^T = \MBF{U}_{s} \sqrt{\MBF{D}_{s}} \left( \MBF{U}_{s} \sqrt{\MBF{D}_{s}} \right)^T = \MBF{C}_{s}\MBF{C}_{s}^T, 
\end{equation*}
where $\MBF{U}_{s}$ and $\MBF{D}_{s}$ are the eigenvector and eigenvalue matrices of $\MBF{B}_{s}$, respectively.
\end{remark}

The reformulation Eq. \eqref{eq:conditional-covariance_reformulation} affords an explicit insight into the conditional covariance operator. As $\MBF{B}_s$ is computed from the gram matrix $\MBF{G}^s_{Y}$, $\MBF{C}_s$ is highly related to the conditional variable $Y$. Compared with the covariance operator on RKHS $\hat{\MBF{R}}^s_{XX}= \MBF{\Phi}_s \MBF{H}_{n}  \MBF{\Phi}_s^T /n$, the feature map $\MBF{\Phi}_s$ in conditional covariance operator $\hat{\MBF{R}}^s_{XX|Y}$ is transformed by the modified centering matrix $\MBF{H}_{n} \MBF{C}_s$ which contains the conditional information. Based on the above reformulation, the following theorem provides the explicit computation of the CKB metric. Note that the reformulation Eq. \eqref{eq:conditional-covariance_reformulation} is included in the proof of Theorem~\ref{thm:empirical-estimation-CKB}, and all proofs of theorems, propositions are provided in the supplementary material.

\begin{theorem}\label{thm:empirical-estimation-CKB}
The empirical estimation of the CKB metric is computed as
\begin{small}
\begin{align}
~& \hat{d}_{\emph{CKB}}^2(\hat{\MBF{R}}^{s}_{XX|Y},\hat{\MBF{R}}^{t}_{XX|Y}) \nonumber \\
=&  \varepsilon {\rm tr} \left[ \MBF{G}^{s}_X \left( \varepsilon n \MBF{I}_n + \MBF{G}^{s}_Y \right)^{-1} \right]
+ \varepsilon {\rm tr} \left[ \MBF{G}^{t}_X \left( \varepsilon m \MBF{I}_m + \MBF{G}^{t}_Y \right)^{-1} \right] \nonumber \\
~& - \frac{2}{\sqrt{nm}} \left\| \left( \MBF{H}_m \MBF{C}_t \right)^T  \MBF{K}^{ts}_{XX}  \left( \MBF{H}_n \MBF{C}_s \right) \right\|_* , \label{eq:empirical-estimation-CKB}
\end{align}
\end{small}
where $\| \cdot \|_*$ is the nuclear norm.
\end{theorem}
\begin{remark}\label{rem:complexity-CKB}
The computational complexity of the CKB metric consists of three terms shown in Eq.~\eqref{eq:empirical-estimation-CKB}. As for the first term, the cost of the kernel matrices and matrix inverse are about $\MC{O}((c+d+n)n^2)$. Similarly, the cost of the second term is about $\MC{O}((c+d+m)m^2)$. As for the third term, the cost of kernel matrix, EVD and nuclear norm is about $\MC{O}(nmd+n^3+m^3+\min(mn^2,m^2n))$. Thus, the computational complexity of the CKB metric is about $\MC{O}(\max(c,d,m,n)(n^2+m^2+mn))$, where $d$ and $c$ are the feature dimension and number of classes, respectively.
\end{remark}

\subsection{Convergence Analysis}

In this section, we focus on the convergence of the empirical estimation of the CKB metric. This convergence theorem is based on the properties of trace-class operator on the Hilbert space and the asymptotic theory of the conditional covariance operator established by Fukumizu \etal \cite{fukumizu2009kernel}. Let $\hat{\MBF{R}}^{(n)}_{XX|Y}$ be the conditional covariance operator drawn i.i.d. from distribution $P_{XY}$ with sample size $n$ which is computed as Eq.~\eqref{eq:conditional-covariance_regular}, Proposition 7 in \cite{fukumizu2009kernel} shows that the estimator $\hat{\MBF{R}}^{(n)}_{XX|Y}$ converges to $\MBF{R}_{XX|Y}$ in probability. Moreover, it shows that the sequence $| \text{tr} (\hat{\MBF{R}}^{(n)}_{XX|Y}) - \text{tr} (\MBF{R}_{XX|Y}) |$ is bounded in probability at rate $\frac{1}{\varepsilon_n \sqrt{n}}$.

With the consistency of conditional covariance operator, we now establish the asymptotic theory for the CKB metric. Assuming that the conditional covariance operators are specified by the source and target domains, we define $n'=\min\{n,m\}$ and the squared CKB metric as $\hat{D}^{(n')}_{\text{CKB}}=\hat{d}_{\text{CKB}}^2(\MBF{R}^{s^{(n)}}_{XX|Y},\MBF{R}^{t^{(m)}}_{XX|Y})$ and $D_{\text{CKB}}=d_{\text{CKB}}^2(\MBF{R}^{s}_{XX|Y},\MBF{R}^{t}_{XX|Y})$. The convergence of $\hat{D}^{(n')}_{\text{CKB}}$ is dominated by the convergence of three terms in Eq.~\eqref{eq:conditional-kernel-bures_define}. Specifically, the convergence of first two terms are concluded from the consistency of conditional covariance operator, and the third term can be deduced to the convergence in trace-norm on the Hilbert space. We present the convergence theorem of the CKB metric as follows.
\begin{theorem}\label{thm:convergence_CKB}
Let the regularization parameter $\varepsilon$ in Eq.~\eqref{eq:conditional-covariance_regular} be a series related to $n'$, \ie, $\varepsilon_{n'}$. Assuming $\varepsilon_{n'}$ satisfies that $\varepsilon_{n'} \to 0$ and $\varepsilon_{n'} \sqrt{n'} \to \infty$ ($n' \to \infty$), then we have
\begin{equation*}\label{eq:convergence_CKB}
|\hat{D}^{(n')}_{\emph{CKB}} - D_{\emph{CKB}}| \to 0 ~~~~ (n' \to \infty)
\end{equation*}
in probability with rate $ (\frac{1}{\varepsilon_n' \sqrt{n'}})^{\frac{1}{2}} $.
\end{theorem}

Theorem \ref{thm:convergence_CKB} shows that the empirical estimation error of the CKB metric converges to 0 as $n\to \infty$ in probability. Specifically, the estimation error $|\hat{D}^{(n')}_{\text{CKB}} - D_{\text{CKB}}|$ is bounded in probability at rate $ (\frac{1}{\varepsilon_n' \sqrt{n'}})^{\frac{1}{2}} $. Compared to the rate $\frac{1}{\varepsilon_n \sqrt{n}}$ of the conditional covariance operator, the square root rate $ (\frac{1}{\varepsilon_n' \sqrt{n'}})^{\frac{1}{2}} $ of the CKB metric comes from the convergence rate of the cross term, \ie, $\MBF{R}^{st}_{XX|Y}$.

\section{Unsupervised Domain Adaptation}
In this section, we tackle the UDA problem by describing the domains as conditional distributions and minimizing the conditional distribution discrepancy under the CKB metric.
\subsection{Conditional Distribution Matching Network}
For UDA problems, $\MC{D}^{s} = \{ (\MBF{x}^s_i,\MBF{y}^s_i) \}_{i=1}^{n}$ is taken as the source domain and $\MC{D}^{t} = \{ \MBF{x}^t_j \}_{j=1}^{m}$ the unlabeled target domain, where $\MBF{x}^{s/t}_i$ represent the observations and $\MBF{y}^{s}_i \in \MBB{R}^{K}$ the one-hot labels with $K$ classes. The primary task is to generalize the classifier $C: \MBF{x} \mapsto \MBF{y}$ trained on both $\MC{D}^{s}$ and $\MC{D}^{t}$ to predict the $\MBF{y}^t_i$. Previous UDA methods assume that the target distribution is shifted from the source distribution (\ie, $P_X^s \neq P_X^t$) and generalize $C$ by minimizing the distribution discrepancy. This assumption only considers the feature distribution, but ignores the discriminant information from the labels. Here we consider the shift of conditional distribution $P_{X|Y}$, which will help the adaptation model to incorporate discriminant information. To learn a conditional distribution matching model, we first design a feature extractor $F$ based on Deep Neural Networks (DNNs), which aims to align the conditional distributions of the domains, \ie, $P_{X|Y}^s$ and $P_{X|Y}^t$. Then the classifier $C: F(\MBF{x}) \mapsto \MBF{y}$ will be trained on the aligned features. Denote the extracted features by $\MBF{Z}^{s/t} = [F(\MBF{x}_1^{s/t}),\ldots,F(\MBF{x}_{n/m}^{s/t})]$ and the soft predictions by $\hat{\MBF{Y}}^{s/t} = [C(\MBF{z}_1^{s/t}),\ldots,C(\MBF{z}_{n/m}^{s/t})]$, where $\sum_{i=1}^{K} \hat{y}^{s/t}_{ij}=1$. The detailed network architecture is provided in the supplementary material.

\begin{figure}
    \centering
    \includegraphics[width=0.46\textwidth]{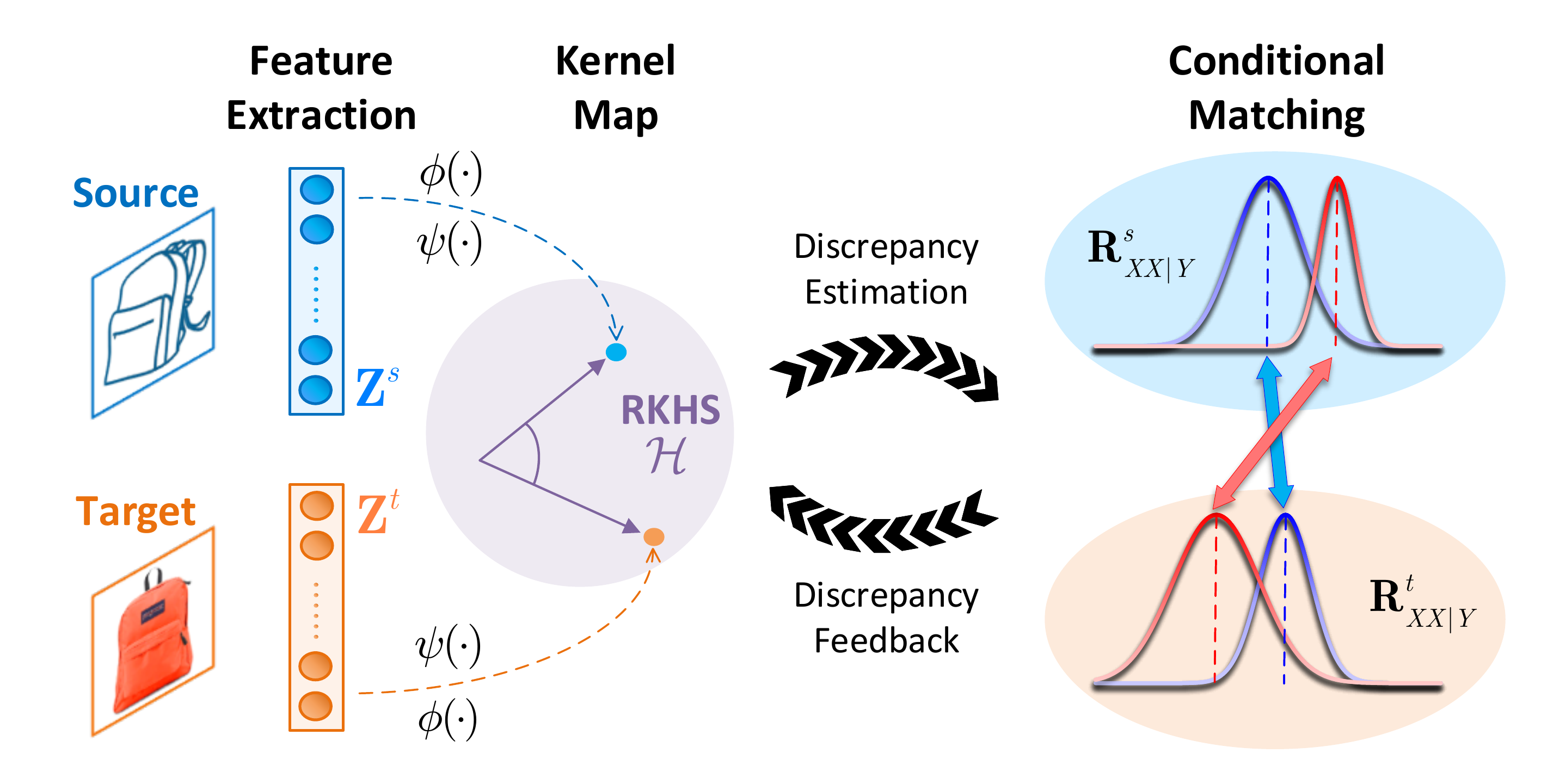}
    \caption{Flowchart of the conditional matching model. The features are mapped into the RKHS, and the conditional distributions of the domains are represented by their conditional covariance operators in RKHS. Then the conditional distribution discrepancy is estimated based on the CKB metric, and the adaptation model is optimized according to the discrepancy feedback.}\label{fig:flowchart_CKB}
    \vspace{-3pt}
\end{figure}

The flowchart of the proposed method is shown in Figure \ref{fig:flowchart_CKB}. It aligns the source and target domains to the conditional invariant space by minimizing the CKB metric between the extracted features, \ie, $\MBF{Z}^{s}$ and $\MBF{Z}^{t}$. Based on the conditional invariant features, a discriminative classifier is learned by applying the entropy-based criterion to both domains. A well-aligned feature space is more preferable for training classifier. Meanwhile, a more accurate classifier leads to a more precise estimation of the CKB metric and fewer misaligned sample pairs. Therefore, the two processes can benefit from each other and enhance the transferability and discriminability of the model alternatively.

In general, the proposed network is trained based on three loss terms. First, the cross-entropy function is applied to the labeled source data, which builds a basic network for classification. The cross-entropy loss $\MC{L}_{\text{CE}}$ is written as
\begin{equation*}\label{eq:loss_CE}
\MC{L}_{\text{CE}} = \sum\nolimits_{i=1}^{K} \sum\nolimits_{j=1}^{n} - y^s_{ij} \log \hat{y}^s_{ij}.
\end{equation*}
Then the entropy $\MC{L}_{\text{Ent}}$ is applied to the target prediction:
\begin{equation*}\label{eq:loss_Ent}
\MC{L}_{\text{Ent}} = \sum\nolimits_{i=1}^{K} \sum\nolimits_{j=1}^{m} - \hat{y}^t_{ij} \log \hat{y}^t_{ij}.
\end{equation*}
This term has been proved to be effective in the semi-supervised learning and unsupervised learning \cite{grandvalet2005semi}. For UDA, it preserves the intrinsic structure of the target domain and reduces the uncertainty of the target prediction.

To match the conditional distributions of two domains, the CKB metric is applied to the deep features learned by the nonlinear mapping $F$. Thus, the kernel matrices $\MBF{K}_{XX}^{s/t},\MBF{K}_{XX}^{ts}$ and feature maps $\MBF{\Phi}_{s/t}$ are computed from the deep features $\MBF{Z}^{s/t}$ hereinafter, \ie, $k(\MBF{z}_i,\MBF{z}_j)$ and $\phi(\MBF{z}_i)$. In terms of the conditional variable $Y$, the kernel matrix $\MBF{K}_{YY}^{s}$ and feature map $\MBF{\Psi}_s$ are computed from the source labels $\MBF{y}_i^s$. As the ground-truth labels $\MBF{y}_i^t$ of the target samples are unknown, we use the pseudo labels $\hat{\MBF{y}}_i^{t}$ to approximate them and compute the feature map as $\hat{\MBF{\Psi}}_t$. The CKB loss is computed according to Eq.~\eqref{eq:empirical-estimation-CKB} as 
\begin{equation*}\label{eq:loss_CKB}
\MC{L}_{\text{CKB}} = \hat{d}^2_{\text{CKB}} (\hat{\MBF{R}}^{s}_{XX|Y},\hat{\MBF{R}}^{t}_{XX|Y}).
\end{equation*}
Let $\lambda_1$ and $\lambda_2$ be the trade-off parameters, the objective function of the conditional alignment model is written as
\begin{equation}\label{eq:objective-CKB}
\min_{F, C} \MC{L}_{\text{CE}} + \lambda_1 \MC{L}_{\text{Ent}} + \lambda_2 \MC{L}_{\text{CKB}}.
\end{equation}

According to Theorem~\ref{thm:conditional-kernel-bures_metricThm}, the domain conditional distributions are aligned (\ie, $P_{X|Y}^s=P_{X|Y}^t$) when $\MC{L}_{\text{CKB}} = 0$. Further, if the marginal distributions $P_{Y}^s$ and $P_{Y}^t$ are also aligned, then the domain joint distribution matching is also achieved as $P_{XY} = P_{X|Y}P_{Y}$. Since the target distribution $P_{Y}^t$ is unknown, we can apply the marginal matching constraint to the label distribution estimated from the classifier's predictions. Specifically, the marginal discrepancy can be approximated by the MMD between $\MBF{\Psi}_s$ and $\tilde{\MBF{\Psi}}_t$, \ie, $\MC{L}_{\text{MMD}} = \| \MBF{\Psi}_s \MBF{1}_n / n - \tilde{\MBF{\Psi}}_t \MBF{1}_m / m \|^2_{\MC{H}_{\MC{Y}}}$, where $\tilde{\MBF{\Psi}}_t$ is computed from the soft predictions $\tilde{\MBF{y}}_i^{t}$. Finally, the joint distribution alignment loss is the sum of $\MC{L}_{\text{MMD}}$ and $\MC{L}_{\text{CKB}}$, and the objective function is written as
\begin{equation}\label{eq:objective-CKB-MMD}
\min_{F, C} \MC{L}_{\text{CE}} + \lambda_1 \MC{L}_{\text{Ent}} + \lambda_2 (\MC{L}_{\text{CKB}} + \MC{L}_{\text{MMD}}).
\end{equation}
In summary, $\MC{L}_{\text{MMD}}$ and $\MC{L}_{\text{CKB}}$ aim to integrate the samples from different domains by mitigating the conditional or joint distribution discrepancies, and the first two terms enhance the model's discriminability by using the label and prediction information from both domains.

\begin{table*}[t]
\setlength{\abovecaptionskip}{0.cm}
\setlength{\belowcaptionskip}{-0.01cm}
\caption{Accuracies (\%) on Office-Home (ResNet-50), Image-CLEF-DA (ResNet-50) and Office10 (AlexNet).}
\label{tab:Office-Home&CLEF&10}
\centering
\renewcommand{\tabcolsep}{0.02pc}
\begin{tabular}{c|cccccccccccc|c}
\toprule[1pt]
\textbf{Office-Home} & Ar$\to$Cl & Ar$\to$Pr & Ar$\to$Rw & Cl$\to$Ar & Cl$\to$Pr & Cl$\to$Rw &
Pr$\to$Ar & Pr$\to$Cl & Pr$\to$Rw & Rw$\to$Ar & Rw$\to$Cl & Rw$\to$Pr & Mean \\
\hline
Source \cite{he2016deep} & 34.9 & 50.0 & 58.0 & 37.4 & 41.9 & 46.2 & 38.5 & 31.2 & 60.4 & 53.9 & 41.2 & 59.9 & 46.1 \\
DAN \cite{long2015learning} & 43.6 & 57.0 & 67.9 & 45.8 & 56.5 & 60.4 & 44.0 & 43.6 & 67.7 & 63.1 & 51.5 & 74.3 & 56.3 \\				
DANN \cite{ganin2016domain} & 45.6 & 59.3 & 70.1 & 47.0 & 58.5 & 60.9 & 46.1 & 43.7 & 68.5 & 63.2 & 51.8 & 76.8 & 57.6 \\
KGOT \cite{zhang2019optimal} & 36.2 & 59.4 & 65.0 & 48.6 & 56.5 & 60.2 & 52.1 & 37.8 & 67.1 & 59.0 & 41.9 & 72.0 & 54.7 \\
CDAN+E \cite{long2018conditional} & 50.7 & 70.6 & 76.0 & 57.6 & 70.0 & 70.0 & 57.4 & 50.9 & 77.3 & 70.9 & 56.7 & 81.6 & 65.8 \\
ETD \cite{li2020Enhanced} & 51.3 & 71.9 & \textbf{85.7} & 57.6 & 69.2 & \textbf{73.7} &  57.8 & 51.2 & \textbf{79.3} & 70.2 & 57.5 & 82.1 & 67.3\\
DMP \cite{luo2020unsupervised} & 52.3 & 73.0 & 77.3 & 64.3 & 72.0 & 71.8 & 63.6 & 52.7 & 78.5 & 72.0 & 57.7 & 81.6 & 68.1 \\
\hline
CKB & \textbf{54.7} & \textbf{74.4} & 77.1 & 63.7 & \textbf{72.2} & 71.8 & 64.1 & 51.7 & 78.4 & \textbf{73.1} & 58.0 & 82.4 & 68.5 \\
CKB+MMD & 54.2 & 74.1 & 77.5 & \textbf{64.6} & \textbf{72.2} & 71.0 & \textbf{64.5} & \textbf{53.4} & 78.7 & 72.6 & \textbf{58.4} & \textbf{82.8} & \textbf{68.7} \\
\bottomrule[1pt]
\end{tabular}
\\[2pt]
\renewcommand{\tabcolsep}{0.58pc}
\begin{tabular}{c|cccccc|c}
\toprule[1pt]
\textbf{Image-CLEF-DA} & I$\to$P & P$\to$I &
I$\to$C & C$\to$I &
C$\to$P & P$\to$C & Mean \\
\hline
Source \cite{he2016deep} & 74.8 $\pm$ 0.3 & 83.9 $\pm$ 0.1 & 91.5 $\pm$ 0.3 & 78.0 $\pm$ 0.2 & 65.5 $\pm$ 0.3 & 91.2 $\pm$ 0.3 & 80.7 \\
DAN \cite{long2015learning} & 74.5 $\pm$ 0.4 & 82.2 $\pm$ 0.2 & 92.8 $\pm$ 0.2 & 86.3 $\pm$ 0.4 & 69.2 $\pm$ 0.4 & 89.8 $\pm$ 0.4 & 82.5 \\				
DANN \cite{ganin2016domain} & 75.0 $\pm$ 0.3 & 86.0 $\pm$ 0.3 & 96.2 $\pm$ 0.4 & 87.0 $\pm$ 0.5 & 74.3 $\pm$ 0.5 & 91.5 $\pm$ 0.6 & 85.0 \\
KGOT \cite{zhang2019optimal} & 76.3 & 83.3 & 93.5 & 87.5 & 74.8 & 89.0 & 84.1 \\
CDAN+E \cite{long2018conditional} & 77.7 $\pm$ 0.3 & 90.7 $\pm$ 0.2 & 97.7 $\pm$ 0.3 & 91.3 $\pm$ 0.3 & 74.2 $\pm$ 0.2 & 94.3 $\pm$ 0.3 & 87.7 \\
ETD \cite{li2020Enhanced} & \textbf{81.0} & 91.7 & \textbf{97.9} & 93.3 & 79.5 & 95.0 & 89.7 \\
DMP \cite{luo2020unsupervised} & 80.7 $\pm$ 0.1 & 92.5 $\pm$ 0.1 & 97.2 $\pm$ 0.1 & 90.5 $\pm$ 0.1 & 77.7 $\pm$ 0.2 & 96.2 $\pm$ 0.2 & 89.1 \\
\hline
CKB & 80.7 $\pm$ 0.1 & \textbf{93.7} $\pm$ 0.1 & 97.0 $\pm$ 0.1 & \textbf{93.5} $\pm$ 0.2 & 79.2 $\pm$ 0.1 & \textbf{97.0} $\pm$ 0.1 & \textbf{90.2} \\
CKB+MMD & 80.7 $\pm$ 0.2 & 92.2 $\pm$ 0.1 & 96.5 $\pm$ 0.1 & 92.2 $\pm$ 0.2 & \textbf{79.9} $\pm$ 0.2 & 96.7 $\pm$ 0.1 & 89.7 \\
\bottomrule[1pt]
\end{tabular}
\\[2pt]
\renewcommand{\tabcolsep}{0.26pc}
\begin{tabular}{c|cccccccccccc|c}
\toprule[1pt]
\textbf{Office10} & A$\to$C & A$\to$D & A$\to$W & C$\to$A & C$\to$D & C$\to$W &
D$\to$A & D$\to$C & D$\to$W & W$\to$A & W$\to$C & W$\to$D & Mean \\
\hline
Source \cite{krizhevsky2012imagenet} & 82.7 & 85.4 & 78.3 & 91.5 & 88.5 & 83.1 & 80.6 & 74.6 & 99.0 & 77.0 & 69.6 & 100.0 & 84.2 \\
GFK \cite{gong2012geodesic} & 78.1 & 84.7 & 76.3 & 89.1 & 88.5 & 80.3 & 89.0 & 78.4 & 99.3 & 83.9 & 76.2 & 100.0 & 85.3 \\
CORAL \cite{sun2016return} & 85.3 & 80.8 & 76.3 & 91.1 & 86.6 & 81.1 & 88.7 & 80.4 & 99.3 & 82.1 & 78.7 & 100.0 & 85.9 \\
OT-IT \cite{courty2016optimal} & 83.3 & 84.1 & 77.3 & 88.7 & 90.5 & 88.5 & 83.3 & 84.0 & 98.3 & 88.9 & 79.1 & 99.4 & 87.1 \\	
KGOT \cite{zhang2019optimal} & 85.7 & 86.6 & 82.4 & 91.4 & 92.4 & 87.1 & 91.8 & \textbf{85.6} & 99.3 & 89.7 & 85.0 & 100.0 & 89.7 \\
DMP \cite{luo2020unsupervised} & 86.6 & 90.4 & \textbf{91.3} & 92.8 & 93.0 & 88.5 & 91.4 & 85.3 & 97.7 & 91.9 & 85.6 & 100.0 & 91.2 \\
\hline
CKB & 87.0 & \textbf{93.6} & 90.2 & \textbf{93.4} & \textbf{93.6} & 90.8 & \textbf{92.7} & 83.5 & \textbf{100.0} & 92.4 & 84.3 & \textbf{100.0} & 91.8 \\
CKB+MMD & \textbf{87.5} & 93.0 & 89.8 & 93.3 & 91.7 & \textbf{92.9} & 92.3 & 83.4 & 99.7 & \textbf{92.8} & \textbf{85.8} & \textbf{100.0} & \textbf{91.9} \\
\bottomrule[1pt]
\end{tabular}
\vspace{-2pt}
\end{table*}

\subsection{Implementation Details}
\vspace{-2pt}
We train the proposed model with back-propagation in the mini-batch manner. As $\MC{L}_{\text{CKB}}$ refers to the inverse of the kernel matrices $\MBF{G}^{s}_Y$ and $\MBF{G}^{t}_Y$, we treat $\hat{\MBF{Y}}^t$ in $\MC{L}_{\text{CKB}}$ as constant to make the optimization stable. Thus, $\MBF{G}^{s}_Y$ and $\MBF{G}^{t}_Y$ are independent of the network parameter and there are no gradients refer to them. The regularization parameter $\epsilon$ of inverse in Eq.~\eqref{eq:empirical-estimation-CKB} is set as $10^{-2}$ empirically. In terms of the kernel function, Gaussian kernel $k(\MBF{x},\MBF{x}')=\exp\left( - \| \MBF{x} - \MBF{x}' \|^2_2 / \sigma^2 \right)$ is adopted, and the parameter ${\sigma^2}$ is set as the mean of the all square Euclidean distances $\| \MBF{x} - \MBF{x}' \|^2_2$ that refer to the corresponding kernel matrix. The
kernel parameters $\sigma$ are adaptively updated for each minibatch. Thanks to the smoothness of the Gaussian kernel, the gradients of the network parameters always exist. The proposed methods in Eq.~\eqref{eq:objective-CKB} and Eq.~\eqref{eq:objective-CKB-MMD} are respectively abbreviated as \textbf{CKB} and \textbf{CKB+MMD} hereinafter. 

\section{Experiment}
\vspace{-2pt}
The proposed methods are evaluated and compared with the SOTA methods on four UDA datasets.


\textbf{ImageCLEF-DA} \cite{caputo2014imageclef} consists of 3 domains with 12 common classes, \ie, \textit{Caltech} (\textbf{C}), \textit{ImageNet} (\textbf{I}), \textit{Pascal} (\textbf{P}), where each domain include 600 images.

\textbf{Office-Home} \cite{OfficeHome} contains 15500 images from 4 domains with 65 classes, \ie, \textit{Art} (\textbf{Ar}), \textit{Clipart} (\textbf{Cl}), \textit{Product} (\textbf{Pr}) and \textit{Real-World} (\textbf{Rw}).

\textbf{Office10} \cite{gong2012geodesic} consists of 4 domains with 10 classes, \ie, \textit{Amazon} (\textbf{A}), \textit{Caltech} (\textbf{C}), \textit{DSLR} (\textbf{D}) and \textit{Webcam} (\textbf{W}).

\textbf{Digits Recognition} Follow the protocol in \cite{hoffman2018cycada}, we conduct the adaptation task between the handwritten digit datasets MNIST (\textbf{M}) and USPS (\textbf{U}).


\subsection{Results}

\textbf{Comparison.} Several state-of-the-art UDA approaches are used to compare with the proposed methods, and the results are shown in Table \ref{tab:Office-Home&CLEF&10}-\ref{tab:Digits}. From the results on Office-Home in Table \ref{tab:Office-Home&CLEF&10}, we observe that the CKB+MMD method outperforms the compared methods in average accuracy, and the relaxed variant CKB also achieves the accuracy of 68.5\%. The experiment results on ImageCLEF-DA are shown in the middle of Table \ref{tab:Office-Home&CLEF&10}. The CKB method improves the mean accuracy to 90.2\% by further considering the discrepancy between the conditional distributions. The results show that the higher the accuracy of target predictions, the more effective the CKB alignment, \eg, tasks \textbf{P $\to$ I} and \textbf{P $\rightarrow$ C}. Table \ref{tab:Office-Home&CLEF&10} shows the results on Office10 dataset. OT-IT and KGOT methods achieve the accuracy of 87.1\% and 89.7\%, which show the superiority of the OT theory in distribution matching. CKB+MMD method achieves Top-1 accuracy in most tasks and improves the mean accuracy to 91.8\%. Table \ref{tab:Digits} shows the results on digits recognition tasks. The proposed models surpass the advanced OT-based method ETD and achieves the highest accuracy in all tasks.

\begin{figure*}[t]
    \begin{minipage}{0.245\linewidth}
        \centering{\includegraphics[width=0.99\linewidth]{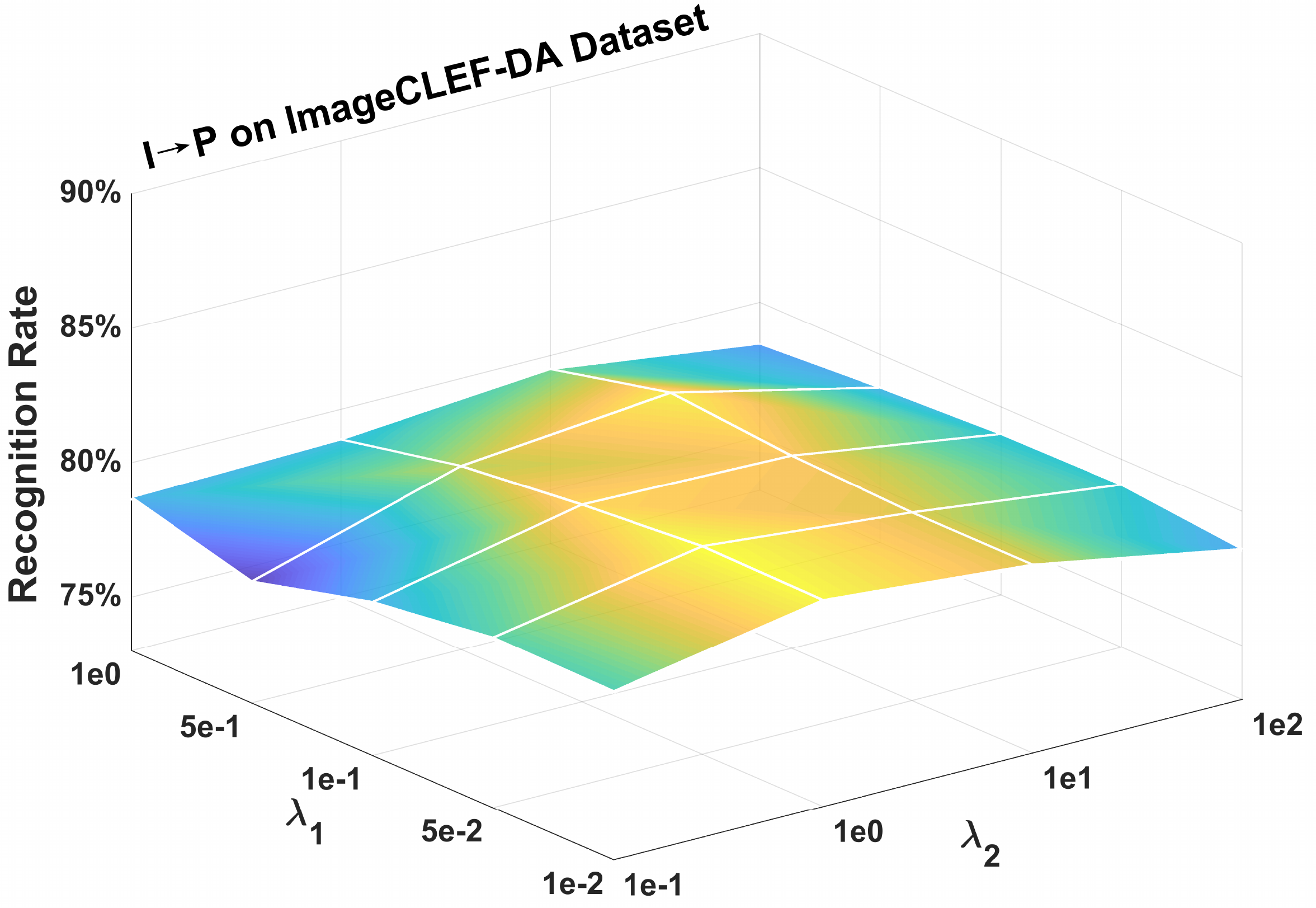}} \\
        (a) Hyper-parameter
    \end{minipage}
    \hfill
    \begin{minipage}{0.245\linewidth}
        \centering{\includegraphics[width=0.99\linewidth]{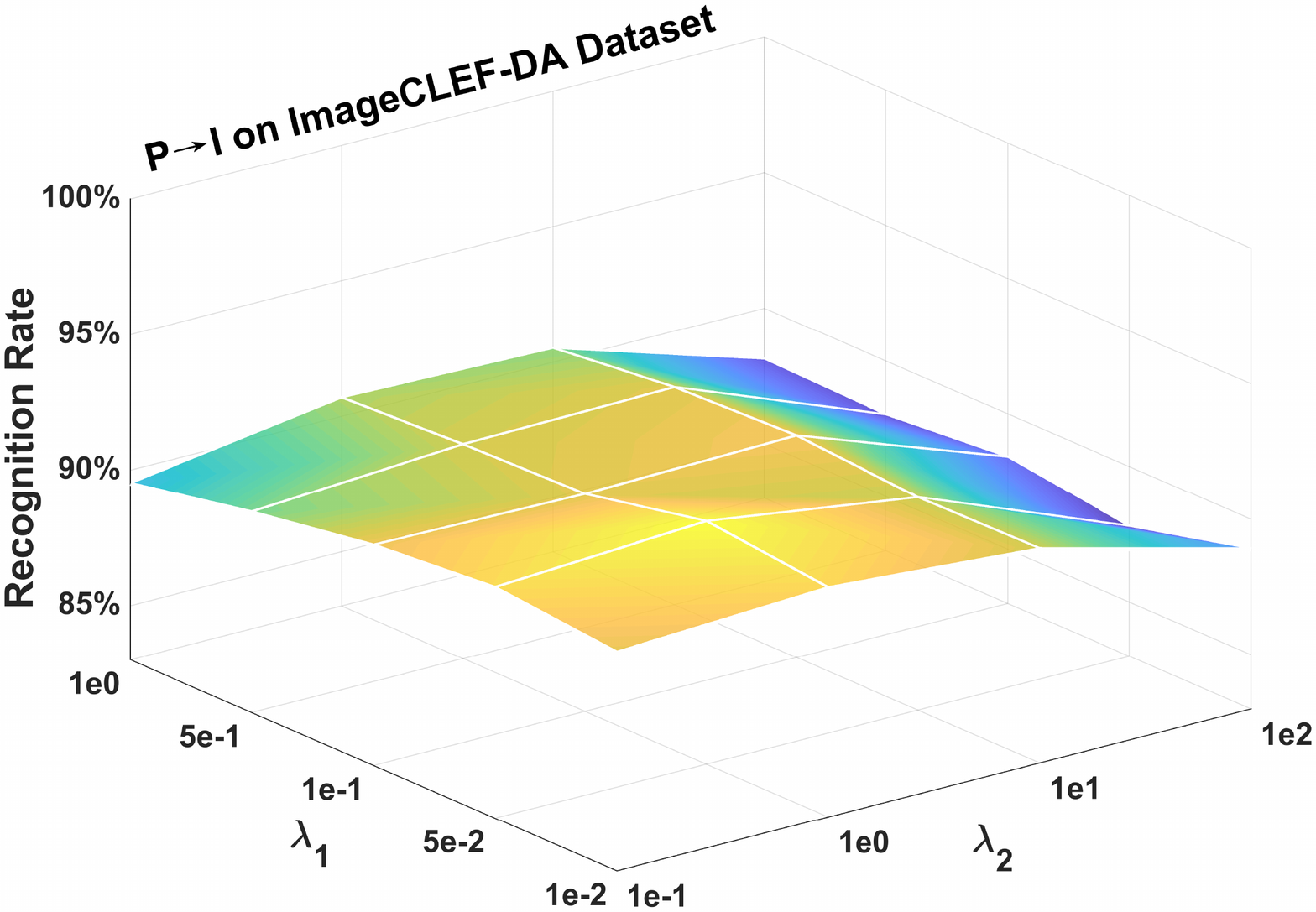}} \\
        (b) Hyper-parameter
    \end{minipage}
    \hfill
    \begin{minipage}{0.245\linewidth}
        \centering{\includegraphics[width=0.99\linewidth]{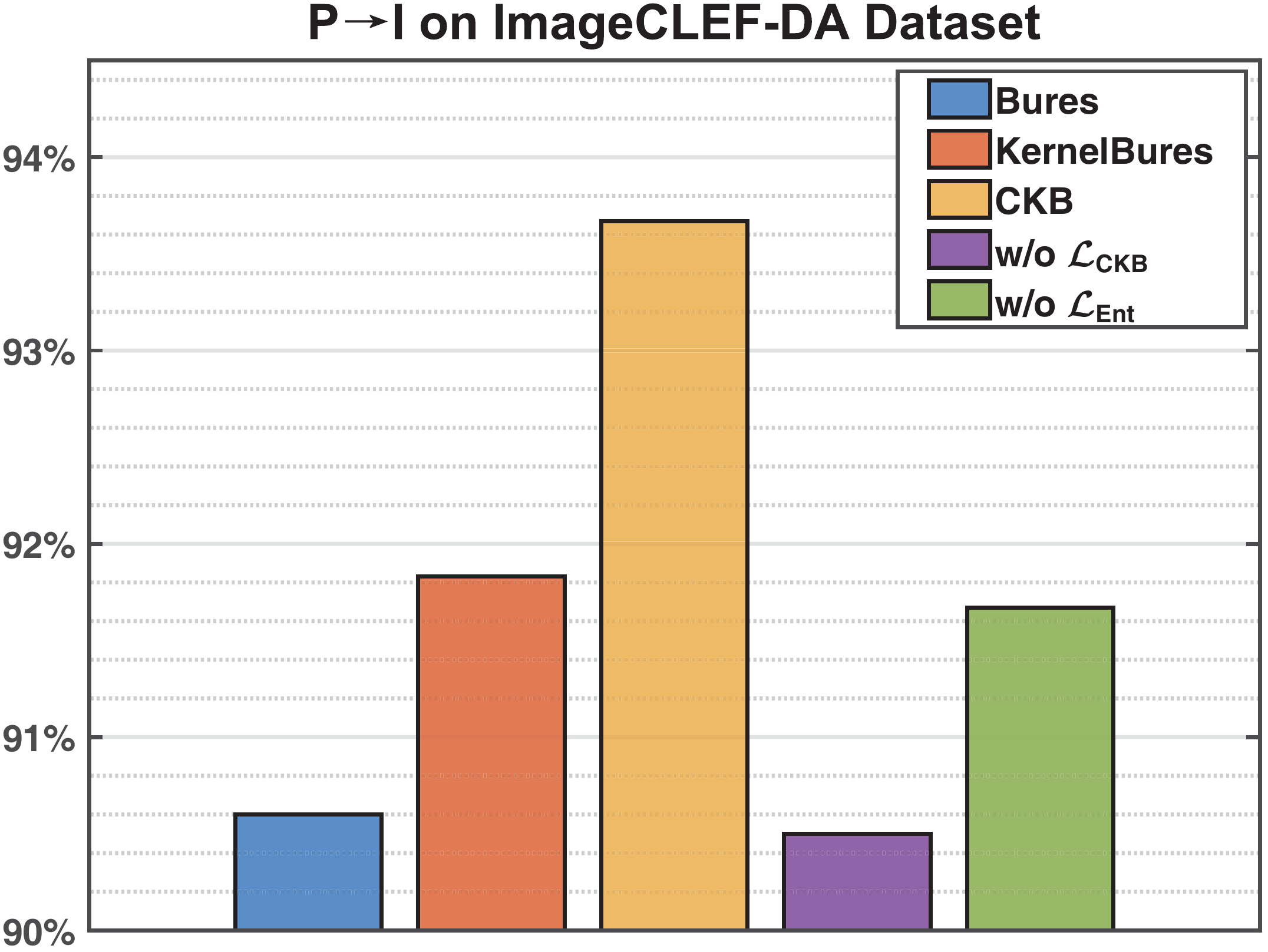}} \\
        (c) Ablation
    \end{minipage}
    \hfill
    \begin{minipage}{0.245\linewidth}
        \centering{\includegraphics[width=0.99\linewidth]{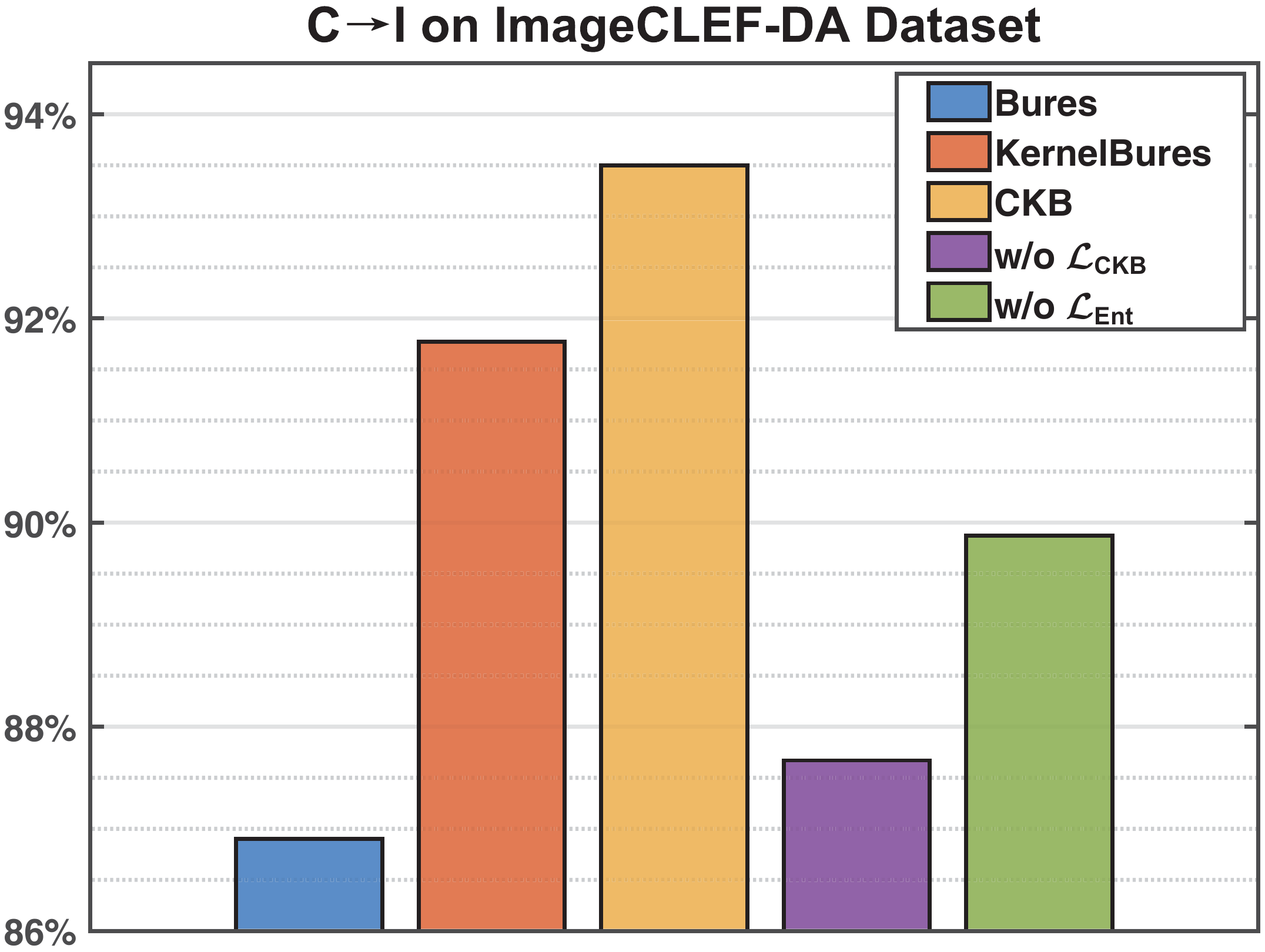}} \\
        (d) Ablation
    \end{minipage}
       \caption{(a)-(b): Grid search for hyper-parameters $\lambda_1$ and $\lambda_2$. (c)-(d): Ablation analysis.}
    \label{fig:Hyper&Ablation}
    \vspace{-1pt}
    \end{figure*}

    \begin{figure*}[t]
    \begin{minipage}{0.245\linewidth}
        \centering{\includegraphics[width=0.99\linewidth]{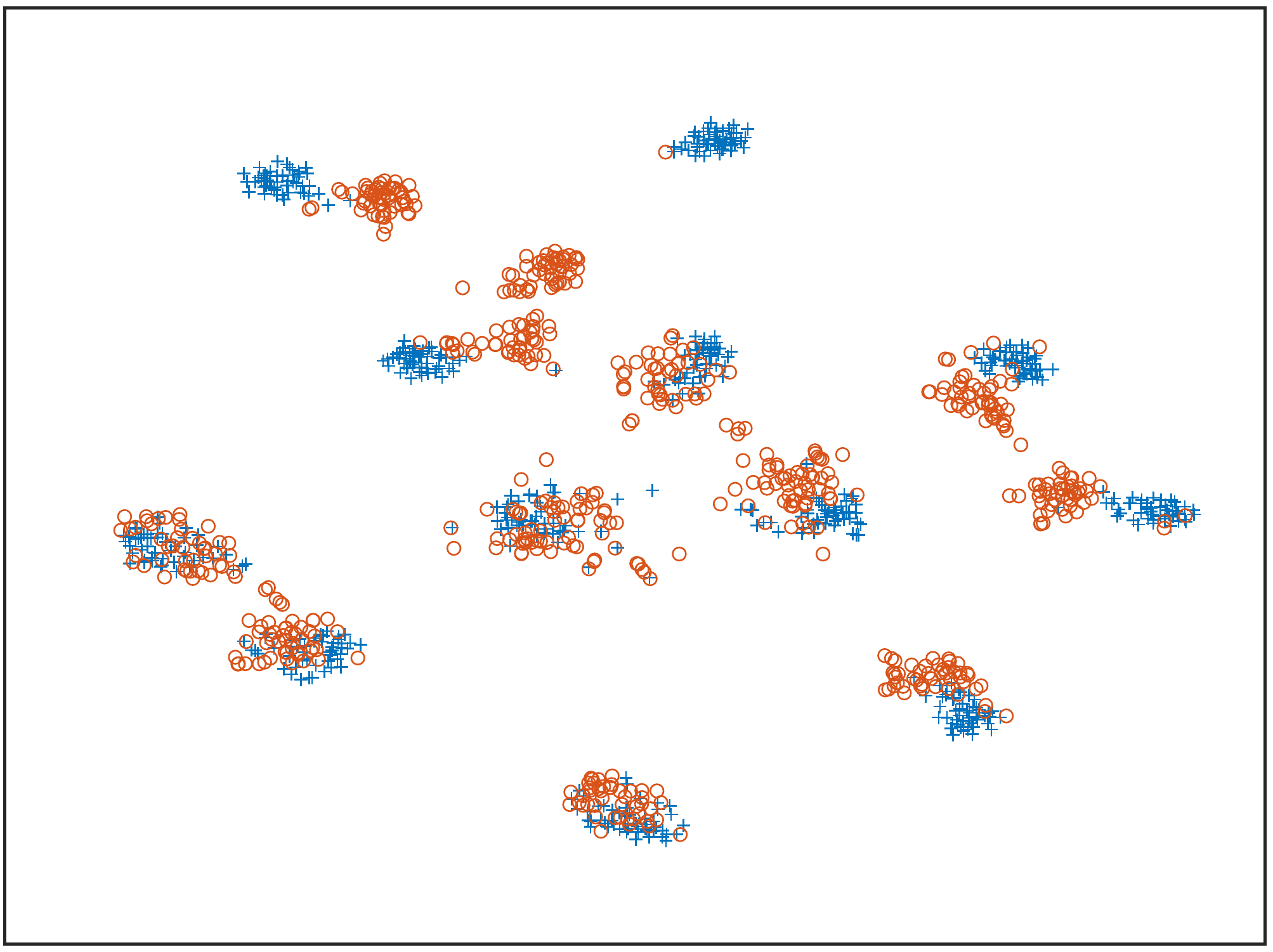}} \\
        (a) Before Adaptation
    \end{minipage}
    \hfill
    \begin{minipage}{0.245\linewidth}
        \centering{\includegraphics[width=0.99\linewidth]{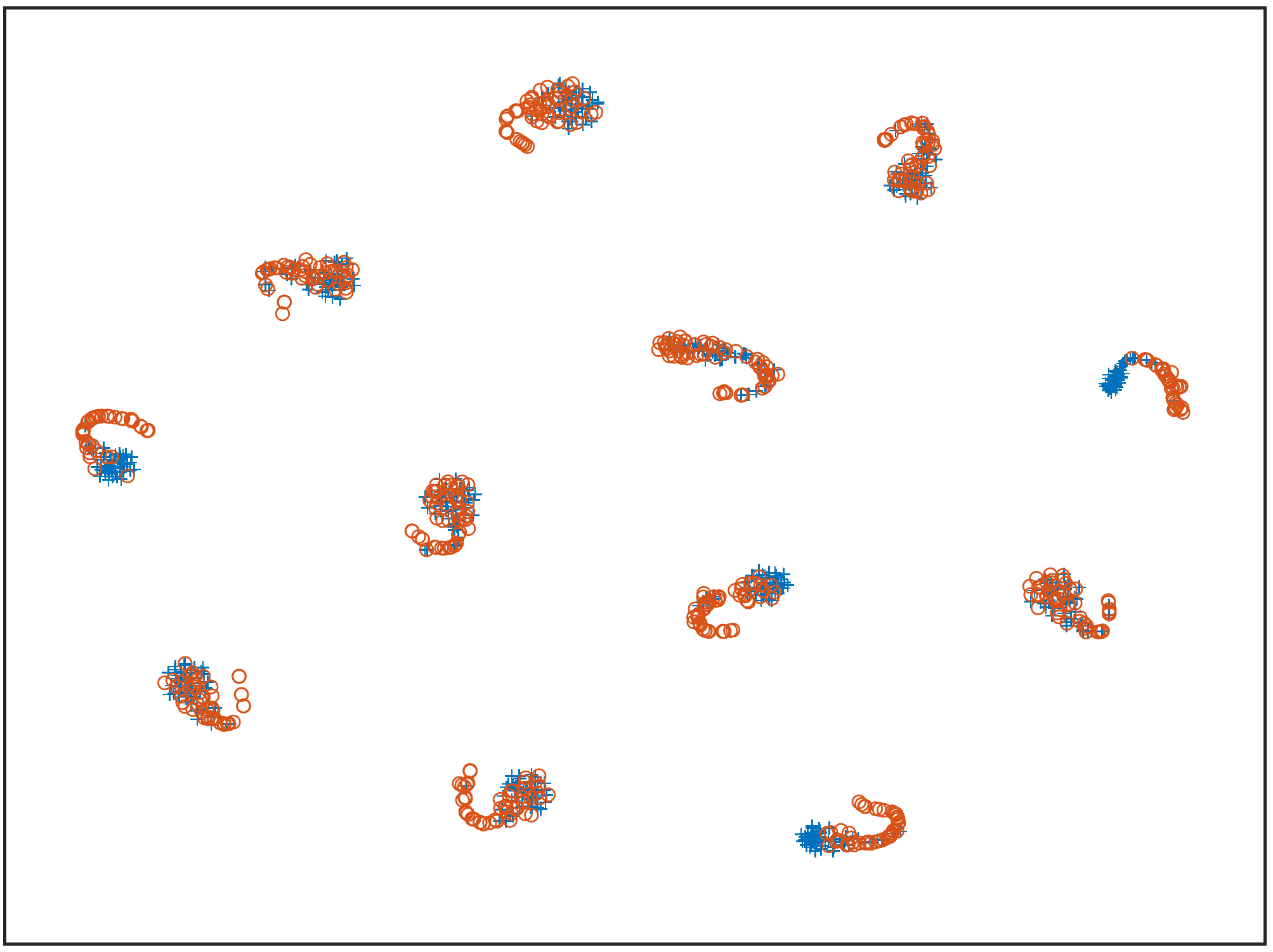}} \\
        (b) After Adaptation
    \end{minipage}
    \hfill
    \begin{minipage}{0.245\linewidth}
        \centering{\includegraphics[width=0.99\linewidth]{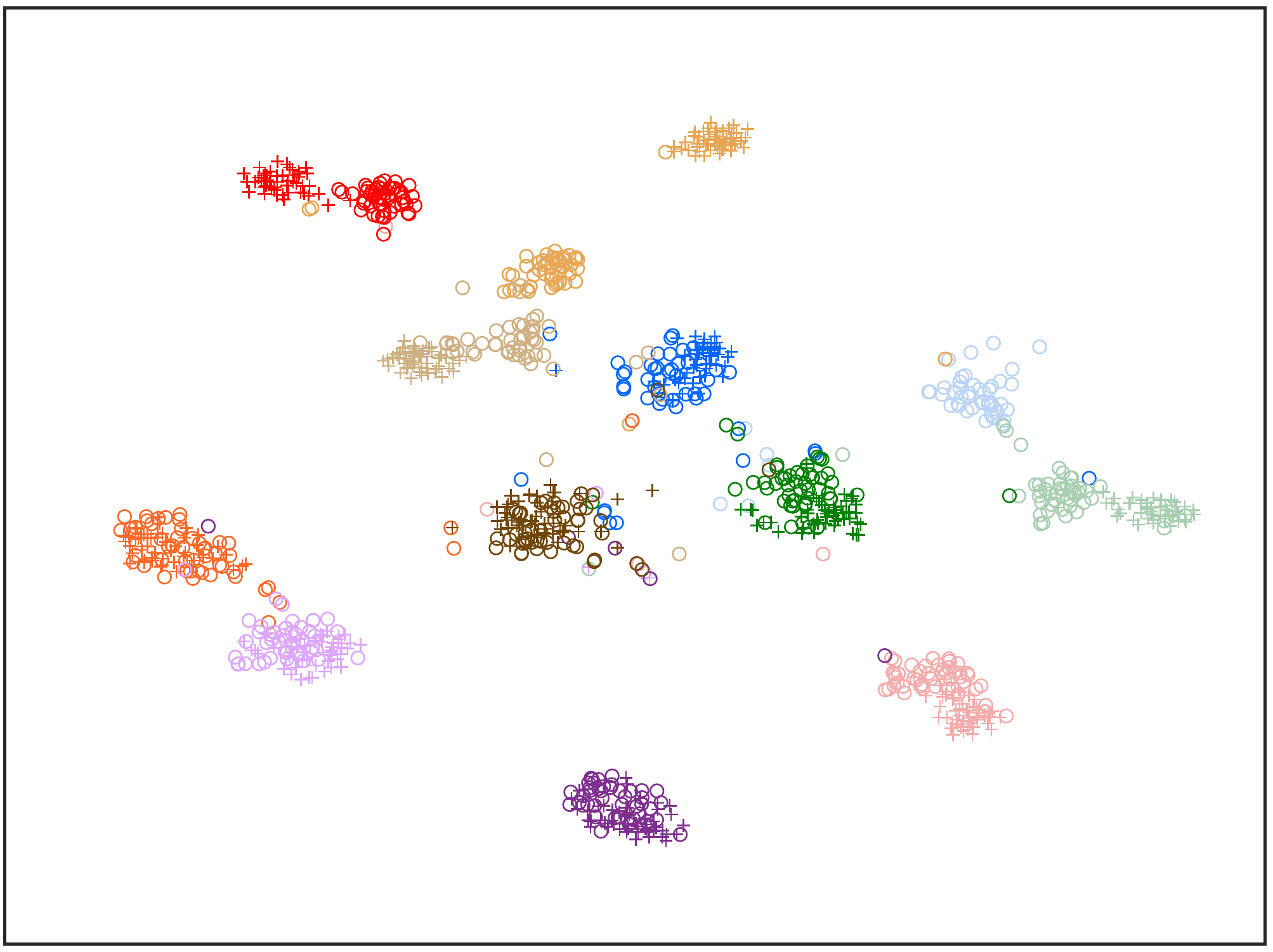}} \\
        (c) Before Adaptation
    \end{minipage}
    \hfill
    \begin{minipage}{0.245\linewidth}
        \centering{\includegraphics[width=0.99\linewidth]{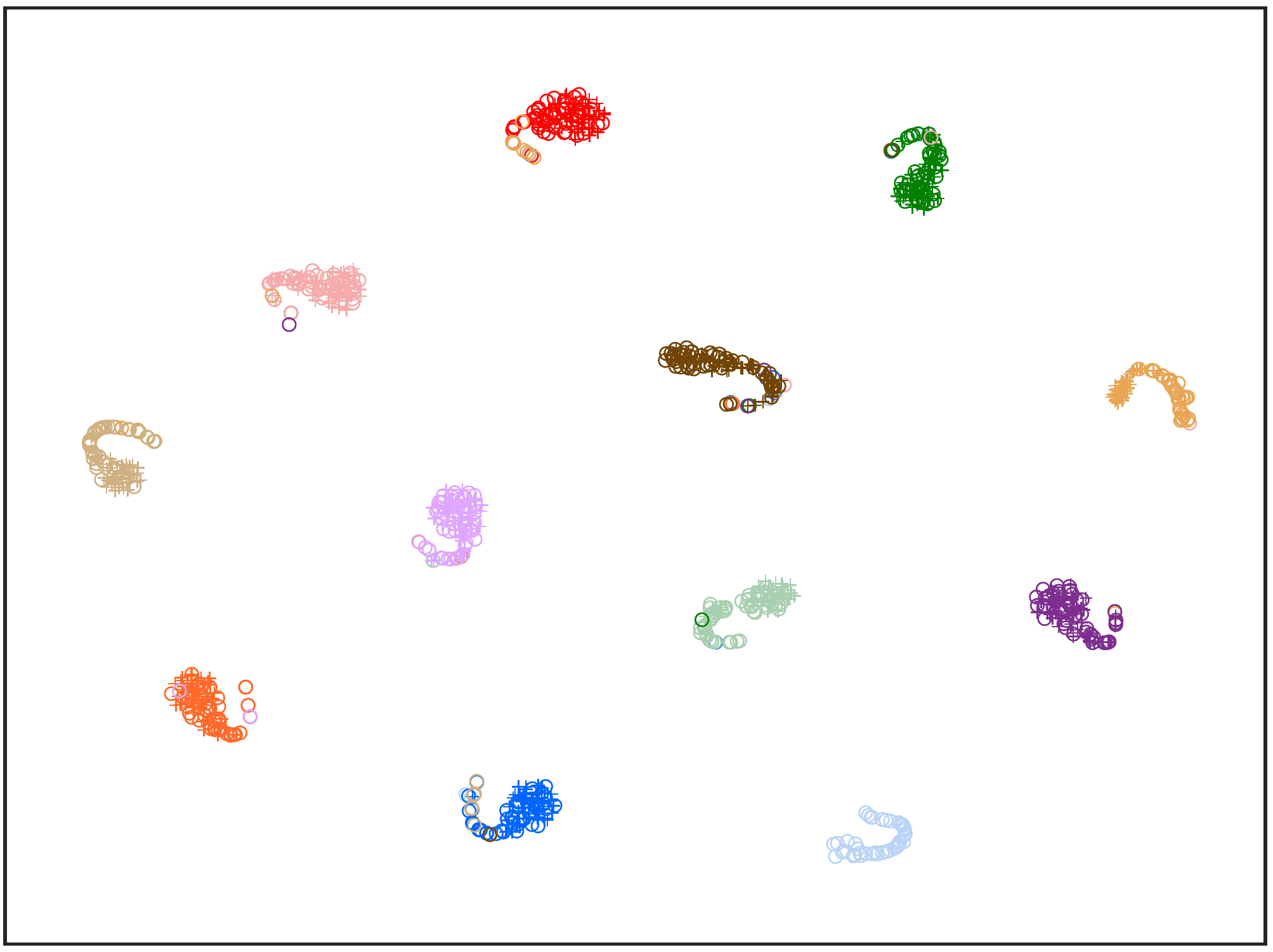}} \\
        (d) After Adaptation
    \end{minipage}
       \caption{Feature visualization of the source-only and CKB models via t-SNE \cite{maaten2008visualizing} on Image-CLEF \textbf{C $\to$ I} task. `+': source domain, `$\circ$': target domain. (a)-(b): Features colored by domains. (c)-(d): Features colored by classes.}
    \label{fig:Tsne}
    \vspace{-5pt}
    \end{figure*}


\begin{table}
\centering
\caption{Accuracies (\%) on Digits (LeNet).}
\label{tab:Digits}
	\begin{tabular}{c|cc}
      \toprule[1pt]
	  Method & M$\to$U & U$\to$M \\
      \hline
      Source \cite{hoffman2018cycada} & 82.2 $\pm$ 0.8 & 69.6 $\pm$ 3.8 \\
      DANN \cite{ganin2016domain} & 95.7 $\pm$ 0.1 & 90.0 $\pm$ 0.2 \\
      CyCADA \cite{hoffman2018cycada} & 95.6 $\pm$ 0.4 & 96.5 $\pm$ 0.2 \\
      DeepJDOT \cite{bhushan2018deepjdot} & 95.7 & 96.4 \\
      ETD \cite{li2020Enhanced} & 96.4 $\pm$ 0.3 & 96.3 $\pm$ 0.1 \\
      \hline
      CKB & 96.3 $\pm$ 0.1 & \textbf{96.6} $\pm$ 0.4 \\
      CKB+MMD & \textbf{96.6} $\pm$ 0.1 & 96.3 $\pm$ 0.1 \\
      \bottomrule[1pt]
	\end{tabular}
\vspace{-4pt}
\end{table}

\textbf{Hyper-parameter.} We investigate the selection of hyper-parameters $\lambda_1$ and $\lambda_2$ on ImageCLEF-DA dataset. The optimal $\lambda_1$ and $\lambda_2$ are respectively searched from [1$e$-2,5$e$-2,1$e$-1,5$e$-1,1$e$0] and [1$e$-1,1$e$0,1$e$1,1$e$2]. Figure \ref{fig:Hyper&Ablation} (a)-(b) show the results of grid search, we observe that the model is stable for different hyper-parameter values and $(\lambda_1,\lambda_2)$=(5$e$-1,1$e$0) is optimal among all settings.

\textbf{Ablation.} We compare the CKB metric with the Bures and Kernel Bures metrics \cite{zhang2019optimal}, and evaluate the effectiveness of the loss terms in Eq.~\eqref{eq:objective-CKB} on ImageCLEF-DA dataset. The model without CKB alignment loss and target entropy loss are abbreviated as w/o $\MC{L}_{\text{CKB}}$ and w/o $\MC{L}_{\text{Ent}}$, respectively. The results in Figure \ref{fig:Hyper&Ablation} (c)-(d) show that the CKB metric is superior to the Bures and Kernel Bures metric, which proves that the conditional operators help the model to obtain the discriminant information from the labels and predictions.


\textbf{Visualization.} To evaluate the aligned features quantitatively, we use t-SNE \cite{maaten2008visualizing} to visualize the features of the source-only model (before adaptation) and the CKB model (after adaptation) on Image-CLEF \textbf{C $\to$ I} task. From Figure \ref{fig:Tsne} (a), we observe that the conditional distribution is still shifted in the source-only model. In Figure \ref{fig:Tsne} (b), all clusters are well-aligned by the CKB method. Figure \ref{fig:Tsne} (c)-(d) show the features colored by classes, we observe that the CKB model achieves the inter-class separability and intra-class compactness on the target domain.

\begin{figure}
\centering
\includegraphics[width=0.45\textwidth]{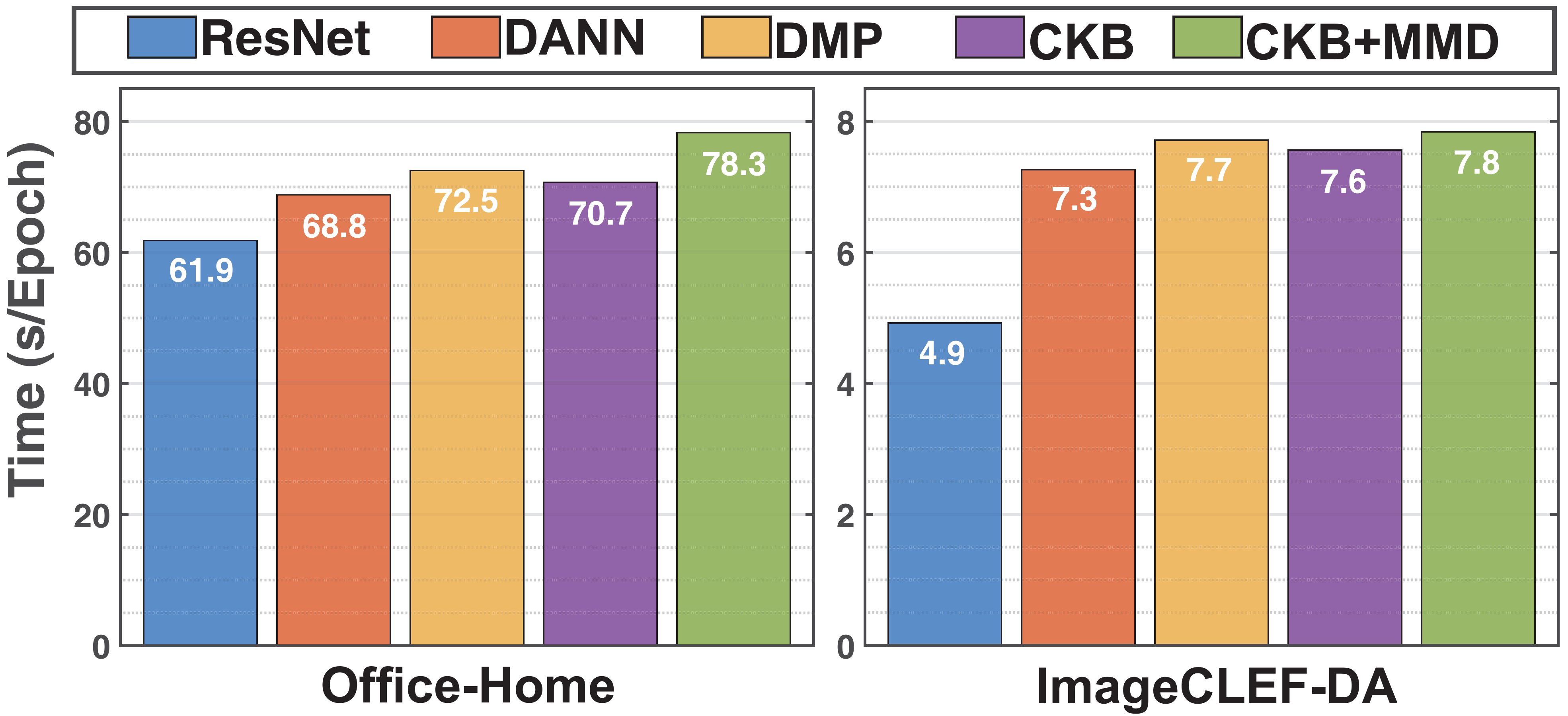}
\caption{Time comparison.}\label{fig:time_comparison}
\vspace{-9pt}
\end{figure}

\textbf{Time Comparison.} We conduct the time comparison experiments on Office-Home and Image-CLEF-DA datasets. The results in Figure \ref{fig:time_comparison} suggest that CKB model is faster than CKB+MMD and DMP, which demonstrates that the conditional discrepancy metric is more efficient than the structure learning model DMP. As the proposed models are trained in mini-batch manner, the time complexity of the CKB metric is only about $\MC{O}(db_s^2)$, where $b_s$ is the batch size. Thus the CKB metric does not introduce much complexity compared to the DNNs. Results show that CKB model only takes 10s longer than ResNet while improving the accuracy significantly by 22\% on Office-Home dataset.

\section{Conclusion}
In this paper, we consider the conditional distribution shift problem in classification. Theoretically, we extend OT in RKHS by introducing the conditional variable, and prove that the proposed CKB metric defines a metric on the conditional distributions. An empirical estimation is derived to provide an explicit computation of the CKB metric, and its asymptotic theory is established for the consistency. By applying the CKB metric to DNNs, we propose a conditional distribution matching network which alleviates the shift of conditional distributions and preserves the intrinsic structures of both domains simultaneously. Extensive experimental results show the superiority of the proposed models in UDA problems.

\section*{Acknowledgement}
This work is supported in part by the National Natural Science Foundation of China under Grants 61976229, 61906046, 11631015, and 12026601.

{\small
\bibliographystyle{ieee_fullname}
\bibliography{references}
}

\end{document}